%% file: main.tex
\documentclass[twoside,11pt]{article}
\pdfoutput=1
\usepackage[nohyperref]{jmlr2e}

\usepackage{times}
\usepackage{epsfig}
\usepackage{graphicx}
\usepackage{amsmath}
\usepackage{amssymb}
\usepackage[pagebackref=true,breaklinks=true,colorlinks,bookmarks=false]{hyperref}
\usepackage{cleveref}

\input{vcl-shortcuts.tex}

\newcommand{\todo}[1]{\textcolor{red}{TODO: #1}}

\definecolor{dgreen}{RGB}{44,160,44}
\definecolor{dyellow}{RGB}{255,127,14}
\definecolor{dred}{RGB}{214,39,40}
\newcommand{\green}[1]{\textcolor{dgreen}{#1}}
\newcommand{\yellow}[1]{\textcolor{dyellow}{#1}}
\newcommand{\red}[1]{\textcolor{dred}{#1}}

\firstpageno{1}

\begin{document}

\title{MINOS: Multimodal Indoor Simulator\\for Navigation in Complex Environments}

\author{\name Manolis Savva
  \email msavva@cs.princeton.edu \\
  \addr Princeton University
  \AND
  \name Angel X. Chang
  \email axchang@cs.princeton.edu  \\
  \addr Princeton University
  \AND
  \name Alexey Dosovitskiy
  \email alexey.dosovitskiy@intel.com \\
  \addr Intel Labs
  \AND
  \name Thomas Funkhouser
  \email funk@cs.princeton.edu  \\
  \addr Princeton University
  \AND
  \name Vladlen Koltun
  \email vladlen.koltun@intel.com \\
  \addr Intel Labs
}


\maketitle

\begin{abstract}
\input{tex/abstract}
\end{abstract}

\input{tex/intro}
\input{tex/related}
\input{tex/simulator}
\input{tex/model}
\input{tex/experiments}
\input{tex/discussion}

{\small
\bibliography{main}
}

\end{document}

%% file: vcl-shortcuts.tex
\usepackage{times}
\usepackage{epsfig}
\usepackage{graphicx}
\usepackage{float}
\usepackage{wrapfig}
\usepackage{amsmath,amssymb} 
\usepackage{algorithm,algorithmicx,algpseudocode}
\usepackage{bm,xspace}
\usepackage{comment}
\usepackage{verbatim}
\usepackage{multirow}
\usepackage{balance}
\usepackage{url}
\usepackage{booktabs}
\usepackage{etoolbox,siunitx}
\usepackage{calc}
\usepackage{pifont,hologo}
\usepackage{color}

\usepackage[super]{nth}
\usepackage{nicefrac}
\robustify\bfseries

\def\bb{\mathbf{b}}

\def\ff{\mathbf{f}}

\def\mm{\mathbf{m}}

\def\oo{\mathbf{o}}
\def\pp{\mathbf{p}}

\def\ss{\mathbf{s}}

\def\aA{\mathcal{A}}

\DeclareMathSymbol{@}{\mathord}{letters}{"3B}

\newcommand\norm[1]{\left\lVert#1\right\rVert}
\newcommand\tuple[1]{\left\langle#1\right\rangle}

\newcommand\timess{\mathbin{\!\times\!}}

\definecolor{alexey}{rgb}{0.8,0.,0.8}
\newcommand\alexey[1]{\textcolor{alexey}{Alexey says: #1}}
\definecolor{manolis}{rgb}{1.0,0.5,0.05}

\definecolor{angel}{rgb}{0.8,0,0.4}

\newcommand\mypara[1]{\noindent\textbf{#1}}

\def\latex/{\LaTeX}
\def\bibtex/{\hologo{BibTeX}}

\usepackage{listings}
\usepackage{inconsolata}  
\usepackage{subcaption}  
\input{js.def}  
\lstdefinestyle{smallnonums}{numbers=none,basicstyle=\fontsize{5pt}{5.25pt}\selectfont\ttfamily}

\usepackage{enumitem}
\newcommand{\denselist}{\itemsep 0pt\parsep=0pt\partopsep 0pt\vspace{-\topsep}}

\newcommand{\unit}[1]{\ensuremath{\,\mathrm{#1}}}
\newcommand{\degree}{\ensuremath{^\circ}\xspace}

\makeatletter
\newcommand*{\centerfloat}{%
  \parindent \z@
  \leftskip \z@ \@plus 1fil \@minus \textwidth
  \rightskip\leftskip
  \parfillskip \z@skip}
\makeatother

%% file: tex/abstract.tex
We present MINOS, a simulator designed to support the development of multisensory models for goal-directed navigation in complex indoor environments. The simulator leverages large datasets of complex 3D environments and supports flexible configuration of multimodal sensor suites. We use MINOS to benchmark deep-learning-based navigation methods, to analyze the influence of environmental complexity on navigation performance, and to carry out a controlled study of multimodality in sensorimotor learning. The experiments show that current deep reinforcement learning approaches fail in large realistic environments. The experiments also indicate that multimodality is beneficial in learning to navigate cluttered scenes. MINOS is released open-source to the research community at \url{http://minosworld.org}.

%% file: tex/intro.tex
\section{Introduction}


Skillful mobile operation in three-dimensional environments has long been posited as an essential milestone on the road to general intelligence~\citep{Moravec1984}.
Despite extensive research, navigation remains a challenging problem.
Classical approaches, based on simultaneous localization and mapping~\citep{Durrant-Whyte2006}, are sensitive to noisy sensory input and changes in the environment.
Recent deep-learning-based methods are potentially more robust, but require extensive training and have only been demonstrated to perform well in simple three-dimensional mazes~\citep{mnih2016asynchronous}.


A key bottleneck for developing and benchmarking approaches to sensorimotor control is the logistical difficulty of operating a mobile agent in the physical world. The physical world is constrained to operate in real time; poor performance can cause breakage that requires repairing or replacing the physical system; and the system may need to be supervised by a human during the learning process. Moreover, in order to ensure proper generalization, a control system must be evaluated in a wide variety of environments.

Due to these limitations, sensorimotor control models are often developed and benchmarked in simulation~\citep{gupta2017cognitive,zhu2016target}.
Once a promising model has been developed and validated, it can be transferred to the physical world~\citep{Pomerleau1988,sadeghi2017cad}.
The transfer is more likely to succeed if the simulator provides a wide range of large and realistic three-dimensional environments.

In this paper, we present MINOS (Multimodal Indoor Simulator)~-- a simulation framework for indoor environments that is designed to support the development and validation of multisensory models for navigation.
MINOS has been designed with several desiderata in mind.
First, the simulator provides access to a large number of realistic environments: the SUNCG dataset of more than $45@000$ three-dimensional models of furnished houses~\citep{song2016ssc} and the Matterport3D dataset of reconstructed indoor scenes~\citep{Chang2017}. Second, the simulator supports flexible multimodal sensing, including vision, depth, surface normals, touch (contact forces), and semantic segmentation. The number of sensors, their positions, and their parameters can be easily specified by the client.
Third, our simulation framework allows for procedural reconfiguration of the environments by programmatic modification of scene composition and appearance.
Finally, the rendering framework is specifically set up to provide high frame-rates -- hundreds of frames per second on a typical workstation -- to support approaches that consume millions of simulation steps during training.


We use MINOS to set up a benchmark for indoor navigation algorithms.
First, we establish fixed train/validation/test splits of varying complexity on both SUNCG and Matterport3D.
This allows for controlled investigation of the generalization of learning-based methods.
Second, we set up three goal-directed navigation tasks: \emph{PointGoal}, \emph{ObjectGoal}, and \emph{RoomGoal}.
The first involves a purely spatial goal specification, while the latter two specify an object type or room type as the goal.
In the \emph{PointGoal} task, the agent is provided with a vector pointing towards the goal;
in the physical world this signal may be provided by an indoor GPS system.
The semantic goal tasks provide the agent with semantic information regarding the goal: a room type (kitchen, bedroom, etc.) or an object type (television, mug, etc.).
This type of command may be provided by a human user interacting with the agent.

Using the presented benchmark, we conduct a controlled study of approaches to sensorimotor learning.
We evaluate several deep reinforcement learning algorithms that navigate towards distal goals using different combinations of sensory modalities.
We find that complex realistic environments present a significant challenge for existing algorithms. For example, in furnished medium-scale Matterport3D scenes, the most successful methods complete the PointGoal task in at most 20\% of trials. The performance on the RoomGoal task is even worse: even in small Matterport3D scenes, the best methods complete the task in only 14\% of trials.

Experiments with varying sensory modalities demonstrate that depth and touch are particularly powerful and can be individually more effective than vision for learning to navigate indoor scenes. Combinations of sensory modalities are more effective still, especially in cluttered environments.

These experiments illustrate the utility of the presented simulation framework for sensorimotor learning research.
To support further research in this direction, MINOS is released open-source to the research community at \url{http://minosworld.org}.

%% file: tex/related.tex
\section{Related Work}


\input{figures_tex/tab_simulators.tex}

Simulation is an established approach to developing, training, and benchmarking sensorimotor control models.
The Arcade Learning Environment~\citep{Bellemare2013} simulates two-dimensional Atari games and has been instrumental in the recent surge of interest in deep reinforcement learning~\citep{mnih2015human,mnih2016asynchronous}.
The ELF platform~\citep{tian2017elf} allows for efficient simulation of 2D real-time strategy games.
Project Malmo enables simulated agents to interface with the game Minecraft~\citep{johnson2016malmo}.
The TORCS~\citep{wymann2000torcs} and CARLA~\citep{Dosovitskiy2017} simulators have been used to study autonomous driving policies.
UAV control has been studied using AirSim~\citep{Shah2017} and UE4Sim~\citep{Mueller2017}.
The Gazebo simulator~\citep{koenig2004gazebo} has been used extensively in robotics research.

VizDoom~\citep{kempka2016vizdoom} and DeepMind Lab~\citep{beattie2016deepmind} simulate stylized immersive three-dimensional labyrinths.
These come close to indoor navigation, but lack realism in terms of layout and appearance, as well as the presence of objects in the scene.
Our work is distinguished from these in its focus on realistic indoor environments.
This allows for development and validation of sensorimotor control models that operate in realistic, cluttered indoor scenes.


Akin to our work, the AI2-THOR project focuses on realistic indoor environments~\citep{zhu2016target}. The goals of AI2-THOR and our work are aligned, but MINOS is distinguished in a number of ways. First, we leverage large datasets, including thousands of furnished houses, with realistic interconnected layouts of up to dozens of rooms each, as opposed to 32 single-room environments provided by THOR.
Second, we focus on flexibility of the agent's sensor suite, both in terms of the available sensors (vision, depth, surface normals, segmentation, touch) and their number and parameters. Third, mindful of the data-hungry nature of many deep RL algorithms, MINOS was developed to run at hundreds of simulation steps per second (rendering tens of millions of frames per day) on typical workstations.

The development of MINOS was initiated in Fall 2016 and the core functionality was completed by June 2017, at which time a version of this paper was submitted for conference publication. Since then, a number of independent efforts have investigated navigation in indoor environments using the SUNCG dataset~\citep{house3d2017,home2017,das2017embodied} and the Matterport3D dataset~\citep{mattersim2017}.
MINOS is distinguished from these in several ways.
First, our simulation framework provides a flexible user API, allowing for (a) environment configuration via object addition/removal and material variation, and (b) fully parameterized placement and specification of multimodal sensor suites with an arbitrary number of sensors.
Second, in addition to the simulator itself, we provide a set of specific benchmark tasks for navigation algorithms.
Third, MINOS supports navigation with both continuous and discrete state spaces in both SUNCG and Matterport3D environments.


We leverage the simulator to study the performance of learning-based navigation agents in cluttered indoor environments.
Recent work on visual navigation used an actor-critic model that discretizes the agent and state space~\citep{zhu2016target} and explored explicit map representations for planning~\citep{gupta2017cognitive}.
Other work uses auxiliary tasks or secondary prediction targets to assist learning~\citep{jaderberg2017reinforcement,mirowski2017learning}.
Direct prediction of future measurements or rewards also appears effective for sensorimotor learning in immersive environments~\citep{dosovitskiy2016learning}.
All these methods have been developed in different environments, which lack in either realism or scale.
Our work provides a fair comparison of a representative set of state-of-the-art deep navigation models in large, complex, and diverse indoor environments.

%% file: figures_tex/tab_simulators.tex

\begin{table*}
\centering
\resizebox{\textwidth}{!}{
\begin{tabular}{@{}lcccccc@{}}
\toprule
Simulator & Agent & Modalities & Framerate & Environment & Dataset size \\
\midrule
Gazebo & \multirow{2}{*}{\green{articulated}} & \green{sensor} & \multirow{2}{*}{\red{10s+ FPS}} & \green{indoor} & \multirow{2}{*}{\red{few environments}} \\
\citep{koenig2004gazebo} & & \green{plugins} & & \green{+ outdoor}& \\
Project Malmo & \green{continuous} & \multirow{2}{*}{\red{color}} & \multirow{2}{*}{\red{10s+ FPS}} & \multirow{2}{*}{\yellow{Minecraft}} & \multirow{2}{*}{\red{few environments}} \\
\citep{johnson2016malmo} & \green{/discrete} & & & & \\
ViZDoom & \multirow{2}{*}{\green{continuous}} & \multirow{2}{*}{\yellow{color, depth, segm}} & \multirow{2}{*}{\green{1000s+ FPS}} & \multirow{2}{*}{\red{stylized mazes}} & \multirow{2}{*}{\red{few mazes}} \\
\citep{kempka2016vizdoom} & & & & & \\
DeepMind Lab & \multirow{2}{*}{\green{continuous}} & \multirow{2}{*}{\yellow{color, depth}} & \multirow{2}{*}{\green{100s+ FPS}} & \multirow{2}{*}{\red{stylized mazes}} & \yellow{few mazes} \\
\citep{beattie2016deepmind} & & & & & \yellow{+ procedural}\\
AI2-THOR & \green{continuous} & \multirow{2}{*}{\red{color}} & \multirow{2}{*}{\green{100s+ FPS}} & \green{indoor} & \multirow{2}{*}{\red{32 rooms}} \\
\citep{zhu2016target} & \green{/discrete} & & & \green{(synthetic)} & \\
CMP & \multirow{2}{*}{\red{discrete}} & \multirow{2}{*}{\yellow{color, depth}} & \multirow{2}{*}{\red{10s+ FPS}} & \green{indoor} & \multirow{2}{*}{\red{6 floors}} \\
\citep{gupta2017cognitive} & & & & \green{(reconstructed)} & \\
CAD2RL & \multirow{2}{*}{\green{continuous}} & \multirow{2}{*}{\yellow{color, depth}} & \multirow{2}{*}{\green{100s+ FPS}} & \yellow{indoor} & \red{12 corridors} \\
\citep{sadeghi2017cad} & & & & \yellow{(synthetic)} & \red{+ variations}\\
\midrule
\multirow{2}{*}{\textbf{MINOS}} & \green{continuous} & \green{reconfigurable} & \multirow{2}{*}{\green{100s+ FPS}} & \green{indoor} & \green{45K houses} \\
 & \green{/discrete} & \green{multimodal} & & \green{(synthetic+reconstructed)} & \green{+ variations} \\
\bottomrule
\end{tabular}
}
\caption{A comparison of MINOS to other simulation environments.
}
\label{tab:simulators}
\end{table*}

%% file: tex/simulator.tex
\section{Simulation Framework}

\input{figures_tex/fig_sim_overview.tex}

MINOS is a flexible, efficient, and customizable framework for simulation of large-scale indoor environments.
\Cref{fig:sim_overview} provides an overview of the framework.
We now describe the components of the system in detail.

\subsection{Simulator API}

The MINOS simulator API is designed to be flexible and easy to use.
A generic dataset layer allows the framework to source environments from the dataset pool.
A flexible configuration API supports:
\begin{enumerate}\denselist
    \item[(a)] \textbf{Environment configuration} by selecting subsets of environments with filters predicated on properties of the house and the objects present, and programmatically creating variations of the original scenes by re-texturing object surfaces in a semantically consistent fashion, as well as removing or replacing furniture;
    \item[(b)] \textbf{Agent control configuration} by choosing between discrete and continuous navigation with a parameterized agent physical model;
    \item[(c)] \textbf{Generic sensor specification} allowing for arbitrary configurations of sensors with custom type, position, orientation, resolution, and encoding. Multiple concurrent sensor streams of each type are supported.
\end{enumerate}

The simulator is implemented in a server-client paradigm.
The WebGL-based server is focused on efficiency, and its instances can be deployed in parallel to servers on any OS.
We offer two client APIs: a Python wrapper designed to support efficient RL, and a web client that is particularly useful for interactive exploration and crowdsourced data collection.
Both the Python and web client APIs communicate with backend instances through a WebSocket layer, allowing for distributed training.

\input{figures_tex/fig_dataset.tex}

\subsection{Environments}

MINOS supports navigation in arbitrary environments. At the time of writing, the simulator provides immediate support for two datasets: the SUNCG dataset of synthetic furnished houses~\citep{song2016ssc} and the Matterport3D dataset of reconstructed real buildings~\citep{Chang2017}. Example scenes are shown in \Cref{fig:dataset}.

The SUNCG dataset provides approximately $45@000$ houses with more than 750K rooms of different types. These models support long-range navigation across layouts that are complex both on the inter-room (interconnected floor plans that require traversing from room to room to reach a goal) and the intra-room scale (rooms are densely furnished and navigation requires maneuvering among the furniture).

The Matterport3D dataset consists of 90 multi-floor residences with approximately $2@000$ annotated room regions. These residences are more realistic than the synthetic SUNCG houses, matching the appearance and composition of real environments more closely. We use Matterport3D as a challenging testbed for RL navigation methods.


\subsection{Agent}

The agent is represented by a cylinder proxy geometry with parameterized height, radius and offset from the ground.  We define a set of control commands that inject linear or angular acceleration: \emph{step forward, step back, turn left, turn right, look up, look down, strafe left}, and \emph{strafe right}.  The web client maps these commands to interactive keyboard control, whereas the RL API receives a set of string identifiers (one for each command) to be applied in a given time step.  Each command is parameterized to allow for scaling of the applied acceleration.

The dynamics of the agent is further parameterized by mass, maximum linear and angular speeds, and coefficient of friction.
These parameters along with the simulation time-step duration can be set to implement continuous navigation agents, or to effectively discretize motion.
For convenient and reproducible experimentation, we provide two pre-configured agents: a discrete controls agent (effectively a discrete space gridworld agent) and a continuous controls agent.

\subsection{Multimodal sensory inputs}
Multimodal perception is crucial for development of sensorimotor skills in animals~\citep{smith2005development} and in artificial systems~\citep{mirowski2017learning}.
To support research on multimodal sensorimotor control, we provide a flexible generic sensory input specification API allowing for any number of sensory inputs in a variety of modalities:

\begin{itemize}[leftmargin=16pt]\denselist
	\item \textbf{Vision:} implemented in WebGL through a real-time rasterization rendering engine.  Supports RGB and grayscale output in arbitrary resolutions.
	\item \textbf{Depth:} extracted from the rasterization depth buffer.  Supports byte or short quantized values, or floating point range in meters, and noise model specification.
	\item \textbf{Surface normals:} per-pixel normals computed from the 3D mesh of the environment.
	\item \textbf{Contact forces:} collision detection of agent proxy geometry against 3D object meshes. Provides collision impulse response forces at the positions of specified contact sensors.
	\item \textbf{Semantic segmentation:} per-pixel category labeling corresponding to fine-grained SUNCG and Matterport3D category hierarchies, as well as per-instance labeling (instance segmentation).
	\item \textbf{Measurements:} agent velocity and acceleration, distance and direction to specified navigation target (Euclidean distance and distance along shortest path), and normalized episode time (fraction of episode time elapsed). These measurements can be used for debugging, visualization, or as modality-agnostic training inputs that can indicate progress towards the goal.
\end{itemize}

\subsection{Customization}

MINOS also provides an API for introducing controlled variation in the environments and for defining a variety of navigation goals and corresponding tasks:

\begin{itemize}[leftmargin=16pt]\denselist
	\item \textbf{Material variation:} textures and colors can be sampled in a semantically consistent way (i.e., respecting the observation frequencies of given material textures and colors for each object instance in the dataset).  The variation can be set to respect the training/validation/test splits so as to ensure that material configurations for particular objects are not shared between splits.  This functionality allows for significant augmentation of synthetic 3D environments. Such randomized retexturing has been used in the work of~\citet{dosovitskiy2016learning} and~\citet{sadeghi2017cad}, and was shown to significantly aid generalization.
	\item \textbf{Object clutter variation:} sets of specified categories of objects can be removed from each environment (e.g., all chairs and all tables).
	\item \textbf{Navigation goal specification:} goals can be specified as arbitrary points in space (randomly sampled or manually placed), with threshold distances for success. Instances of an object category or a room category can also be specified as goals. More specifically, any instance of a category, a randomly selected instance, or the closest instance to the agent can be defined as the goal.
	\item \textbf{Task specification:} the task to be performed by the agent is specified through an arbitrary Python function that computes reward signals and episode success or failure given the agent's current and past observations, measurements, and state.  For our experiments we implement the navigate to X task as a distance check between the current agent position and the closest point in the goal region (which is a point, an object, or a room).
\end{itemize}

%% file: figures_tex/fig_sim_overview.tex
\begin{figure*}[t]
\includegraphics[width=\textwidth]{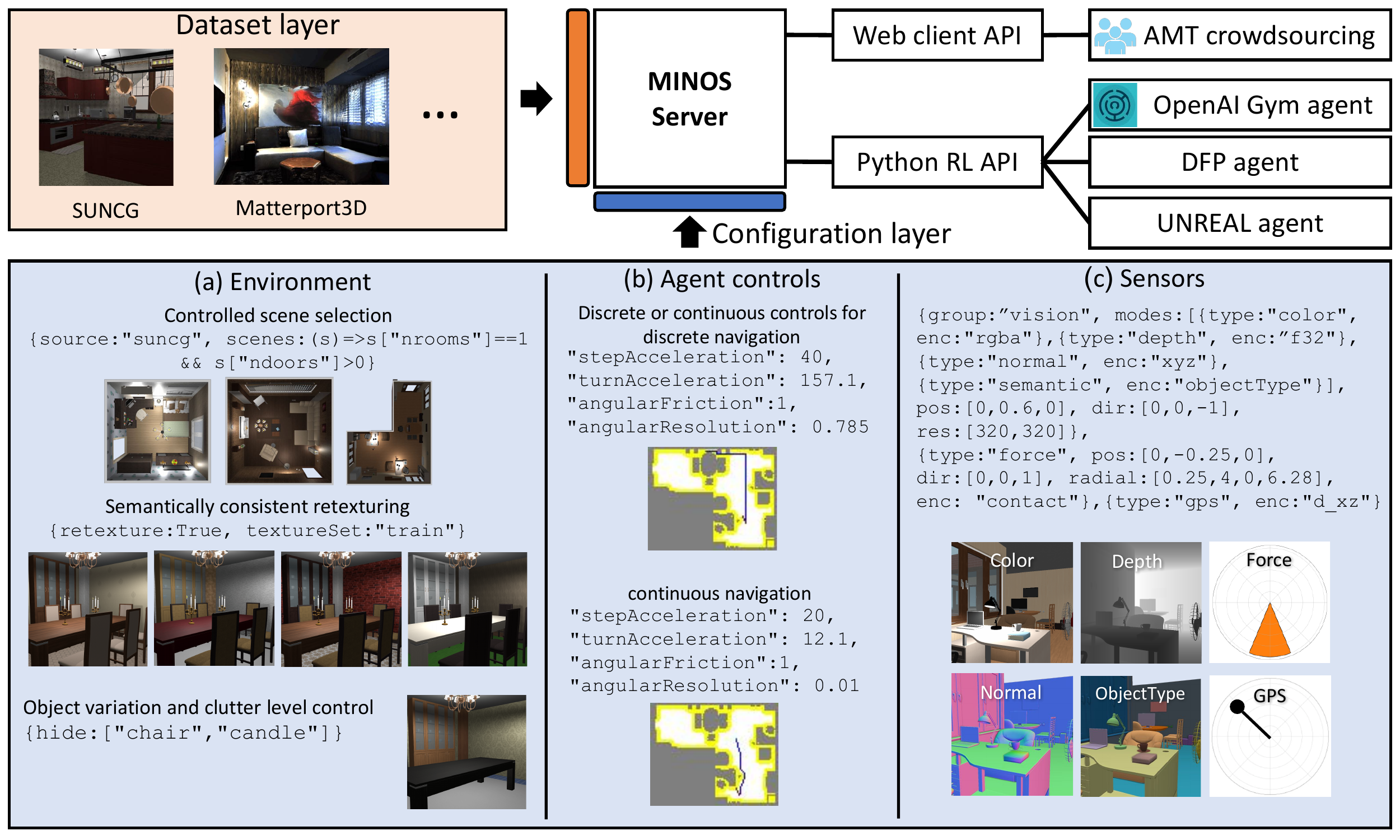}
\caption{Overview of the MINOS framework and APIs. Our framework can source environments from datasets such as SUNCG and Matterport3D, and is accessible both through an RL API and a web client API. (a) Environment configuration. The scripts shown here select all single-room scenes in the SUNCG database for training, enable semantically consistent retexturing, and remove chair and candle objects. (b) Agent controls configuration, with adjustable discrete and continuous navigation parameters resulting in the demonstrated agent trajectories. (c) Agent sensor configuration, which specifies a set of vision, depth, normal, semantic, contact force, and GPS sensors.}
\label{fig:sim_overview}
\end{figure*}

%% file: figures_tex/fig_dataset.tex

\begin{figure*}[t]
\centering
\begin{tabular}{@{}c@{\hspace{0.75mm}}c@{\hspace{2mm}}c@{\hspace{0.75mm}}c@{}}
\includegraphics[width=.24\textwidth,angle=180,origin=c]{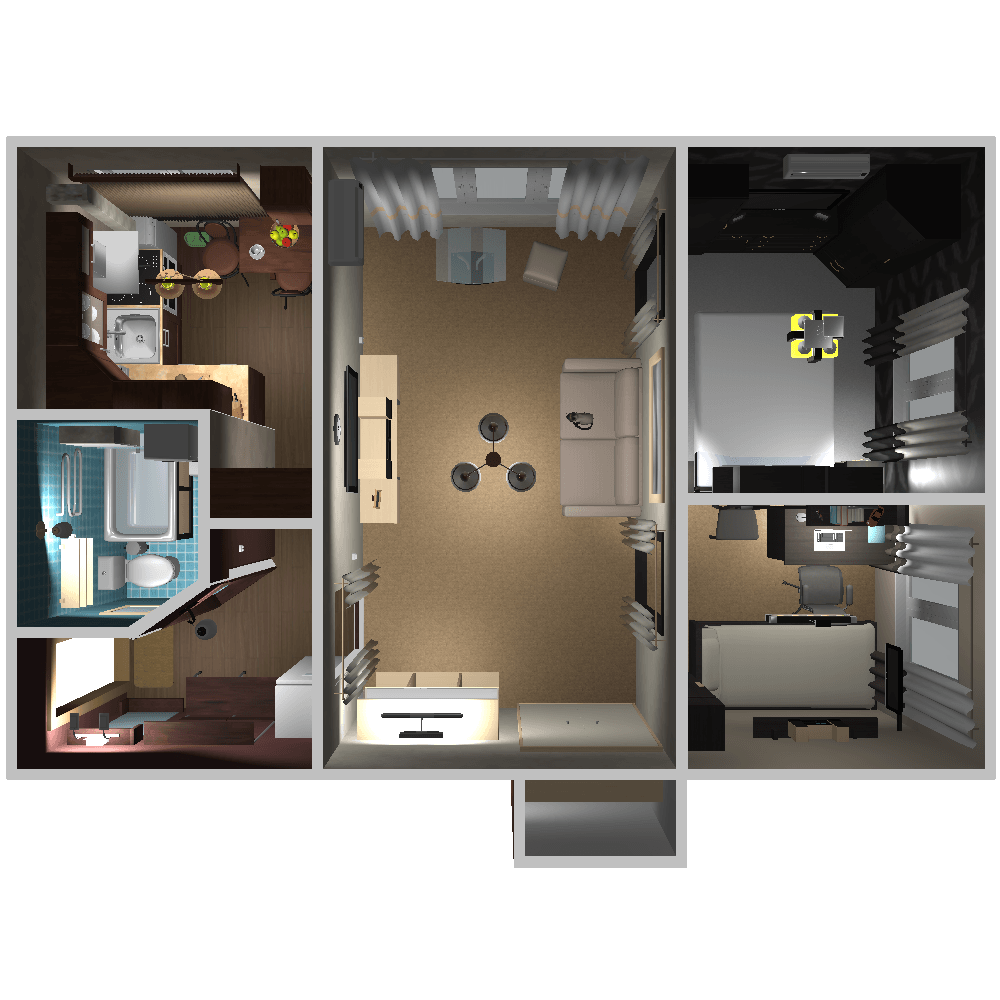}&
\includegraphics[width=.24\textwidth,angle=180,origin=c]{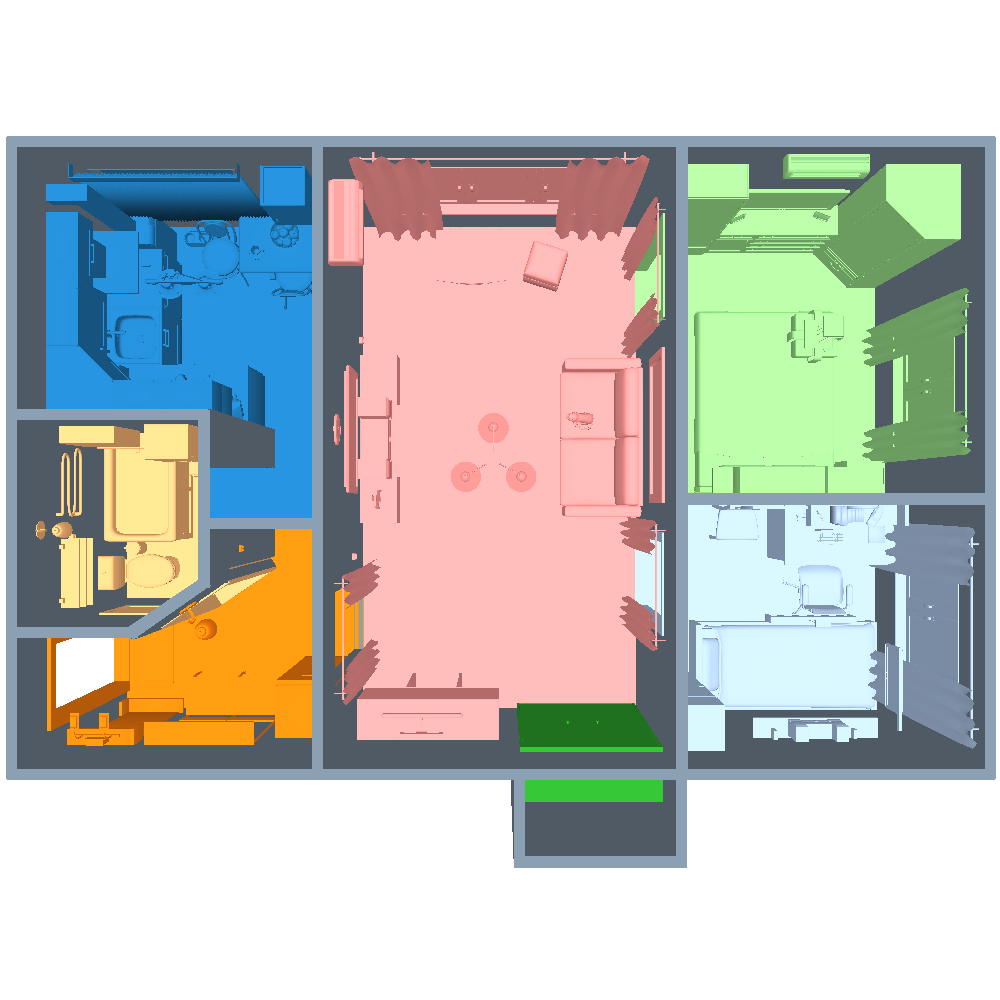}&
\includegraphics[width=.24\textwidth]{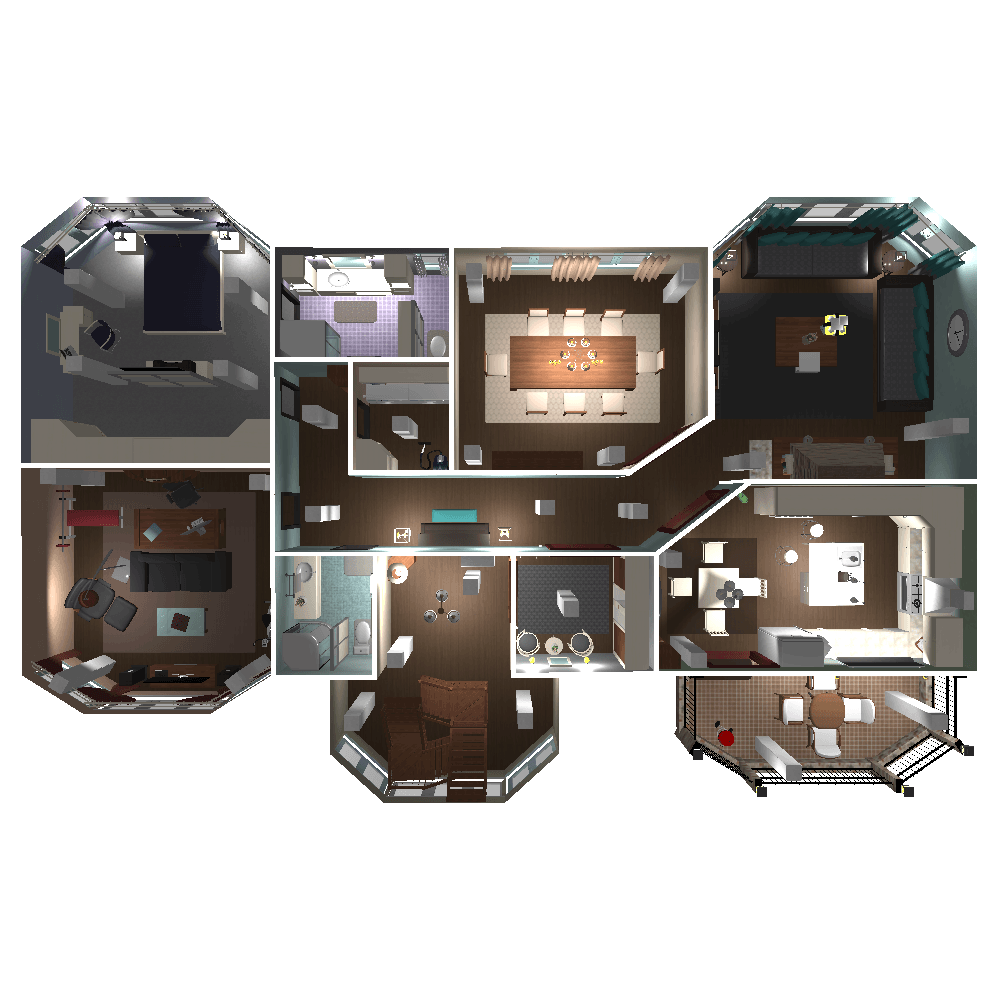}&
\includegraphics[width=.24\textwidth]{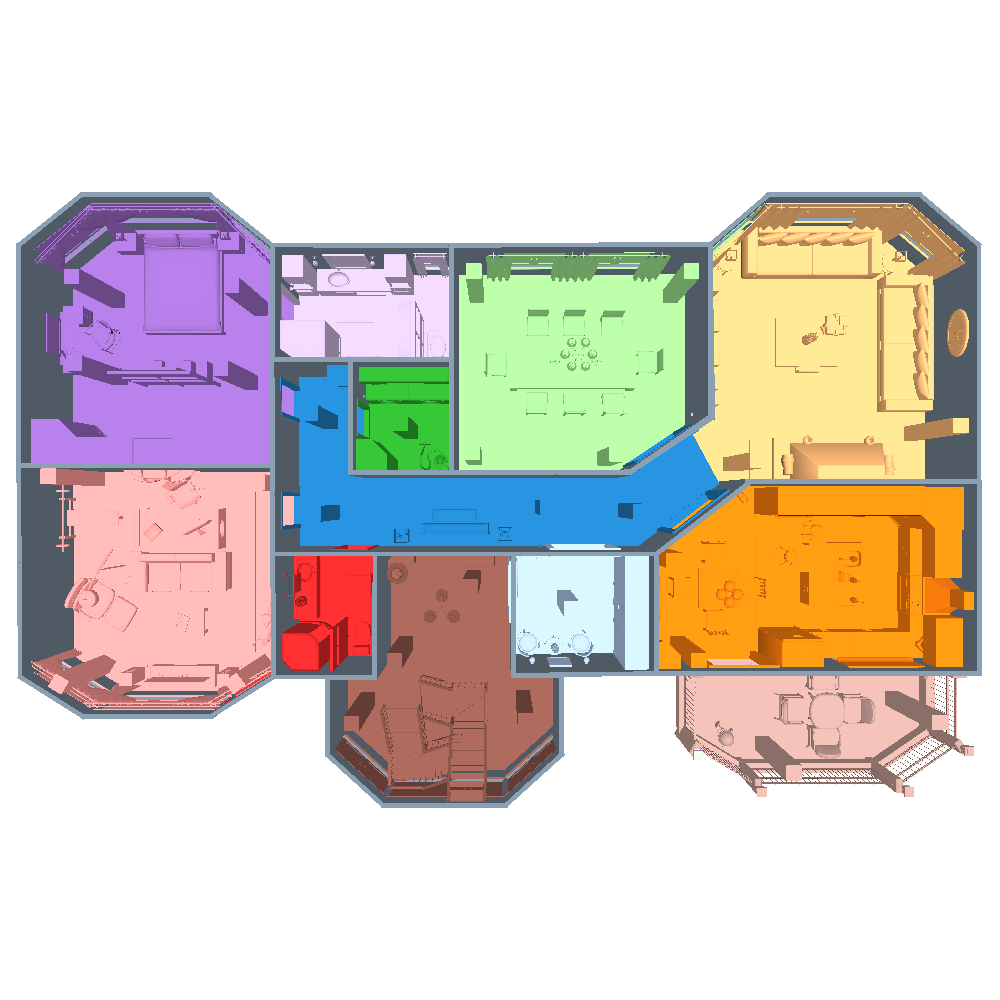}\vspace{-4mm}\\
\includegraphics[width=.24\textwidth]{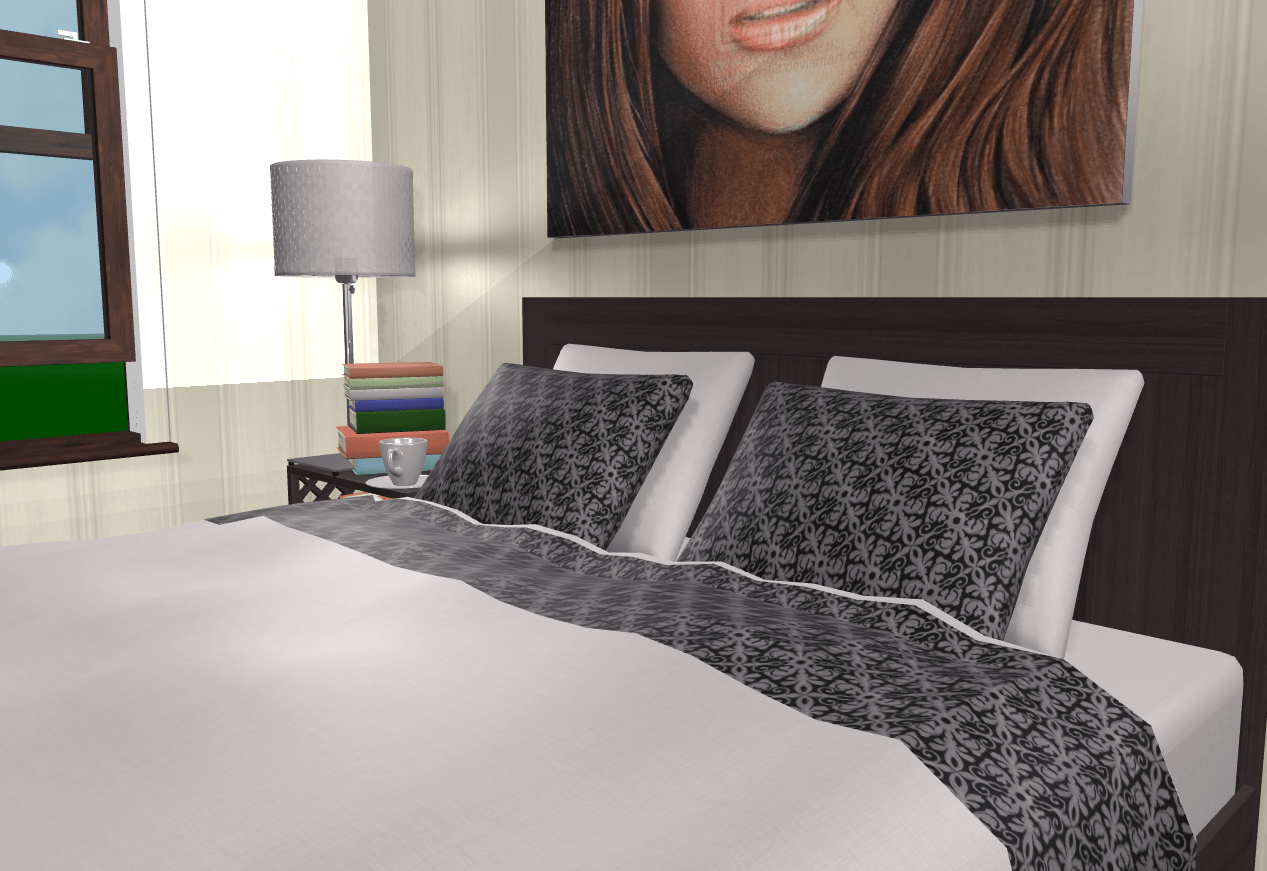}&
\includegraphics[width=.24\textwidth]{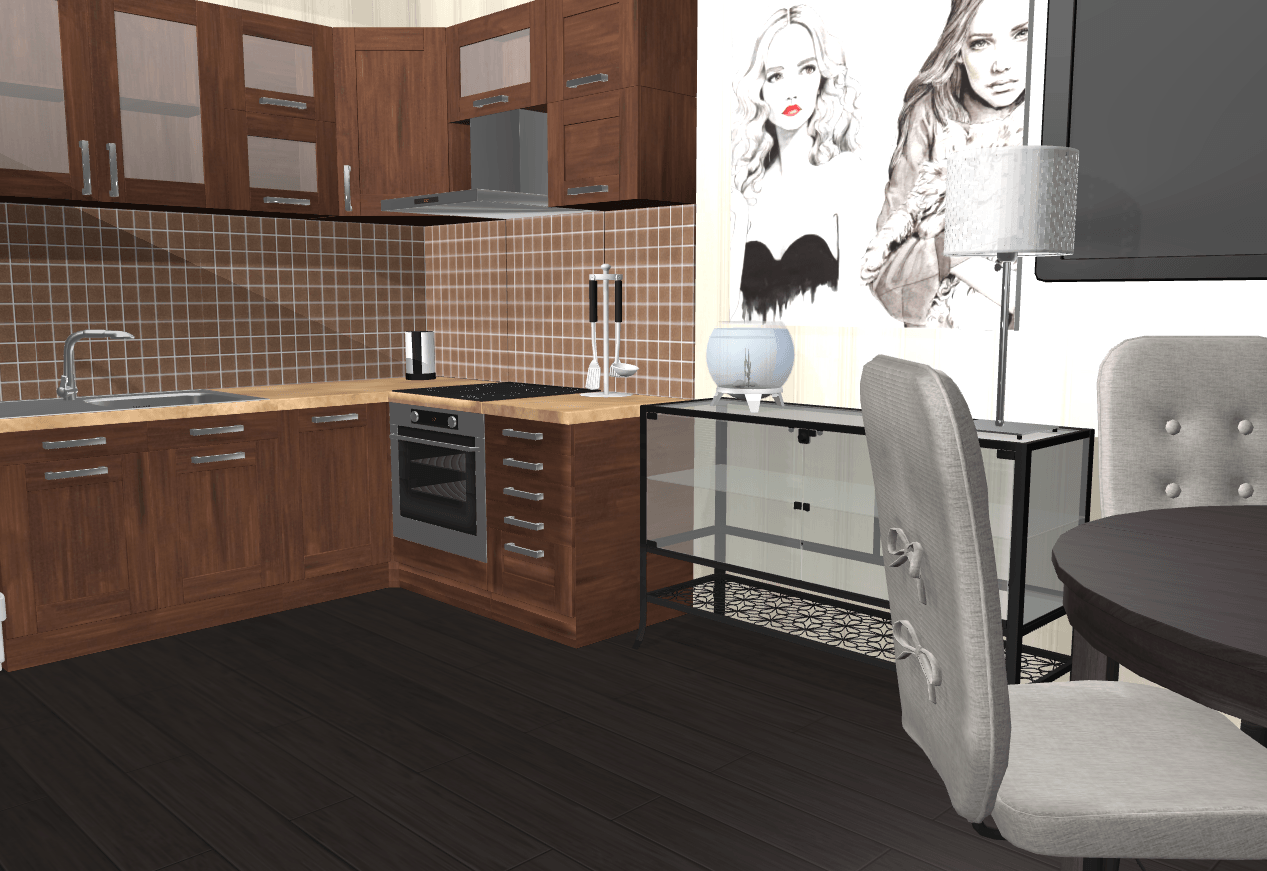}&
\includegraphics[width=.24\textwidth]{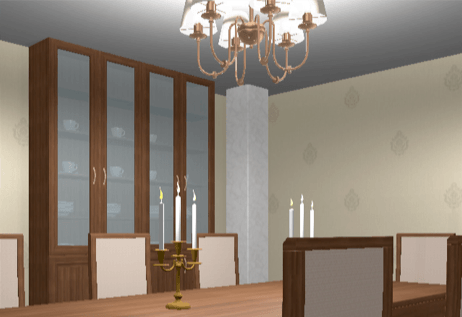}&
\includegraphics[width=.24\textwidth]{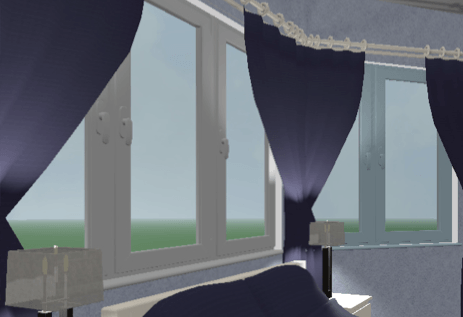}\vspace{-1mm}\\
\multicolumn{2}{c}{(a) 5-room SUNCG house} & \multicolumn{2}{c}{(b) 10-room SUNCG house}\vspace{4mm}\\
\includegraphics[width=.12\textwidth,trim=0 0 5cm 0,angle=90,origin=c]{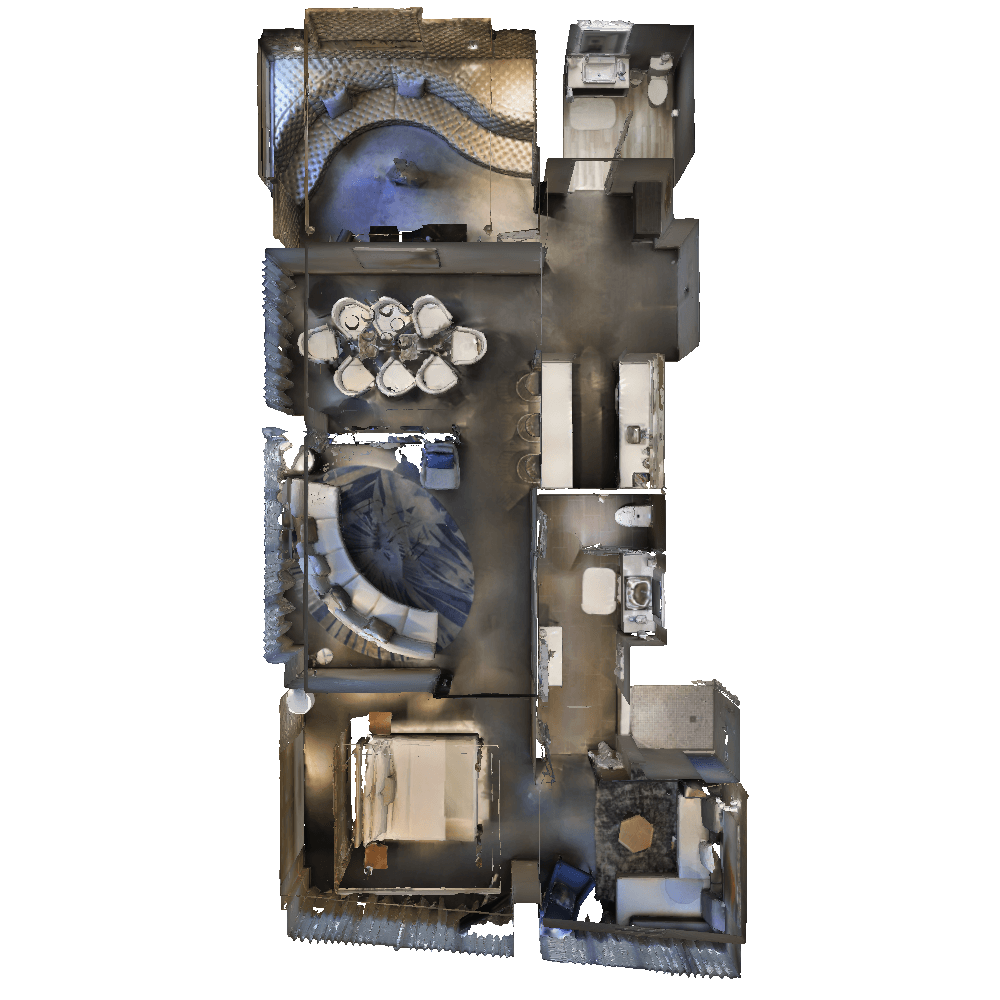}&
\includegraphics[width=.032\textwidth,trim=0 0 8cm 0,angle=90,origin=c]{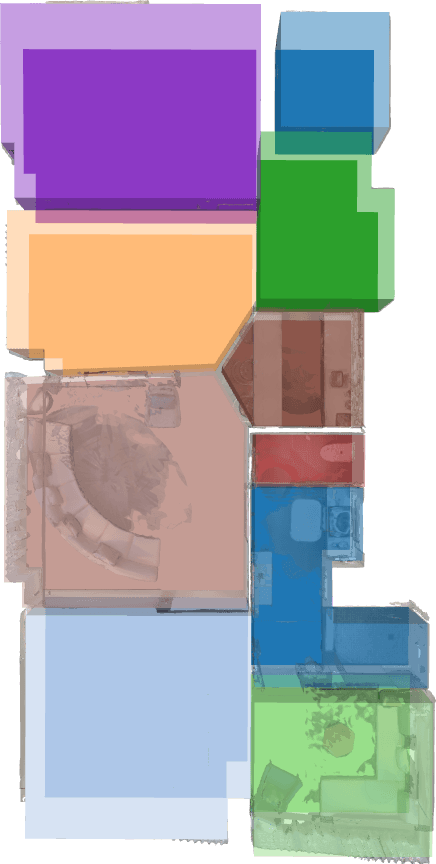}&
\includegraphics[width=.12\textwidth,trim=0 0 5cm 0,angle=90,origin=c]{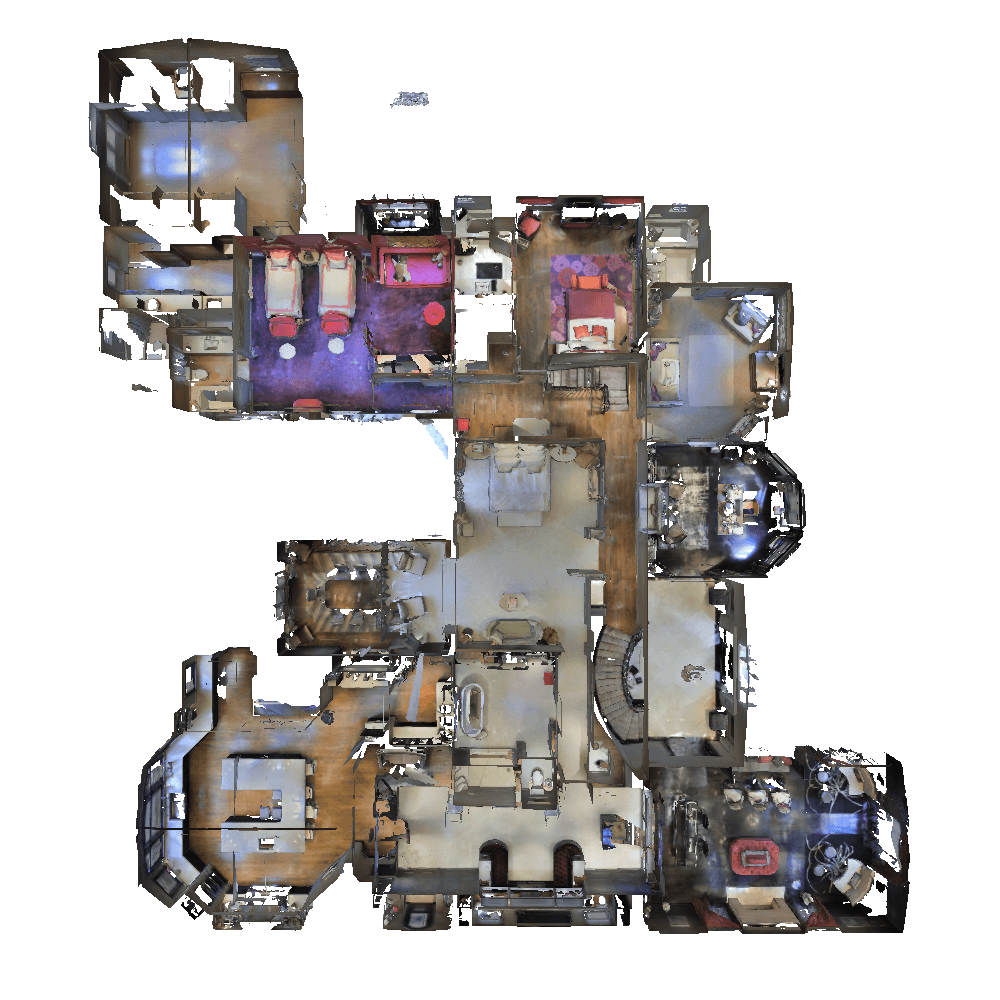}&
\includegraphics[width=.085\textwidth,trim=0 0 8cm 0,angle=90,origin=c]{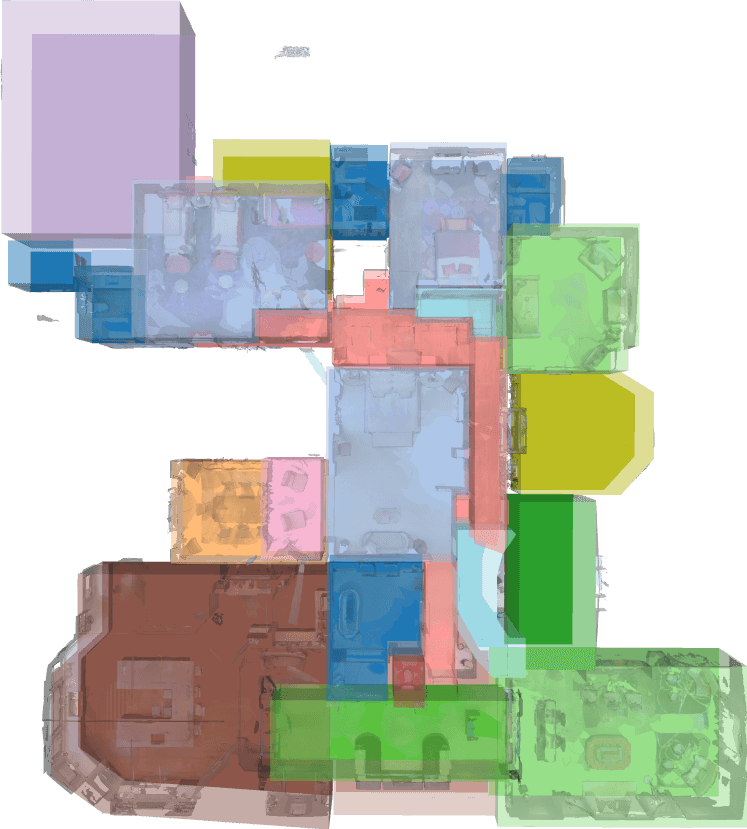}\vspace{-4mm}\\
\includegraphics[width=.24\textwidth,trim=0 0 0 2cm,clip]{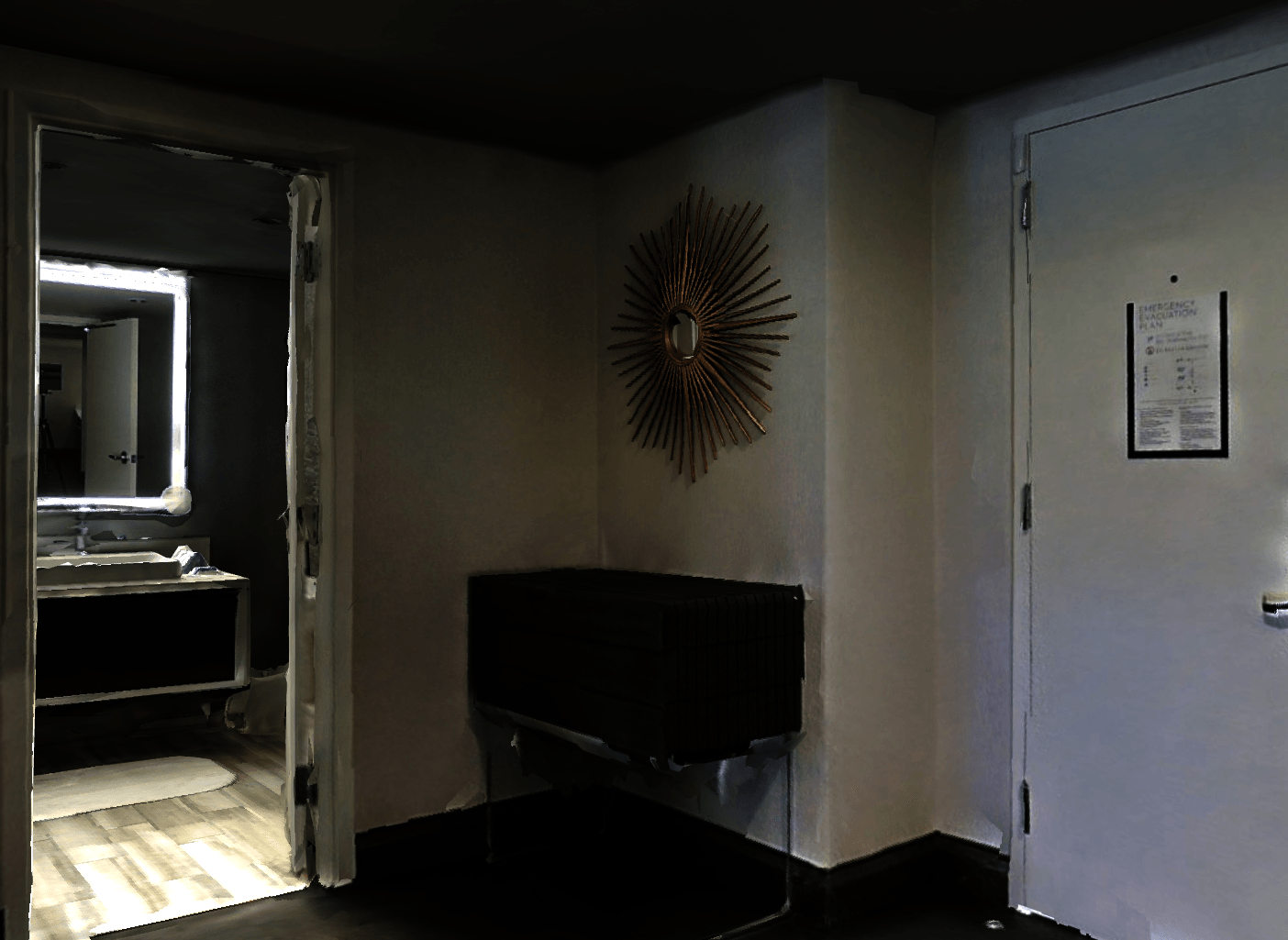}&
\includegraphics[width=.24\textwidth]{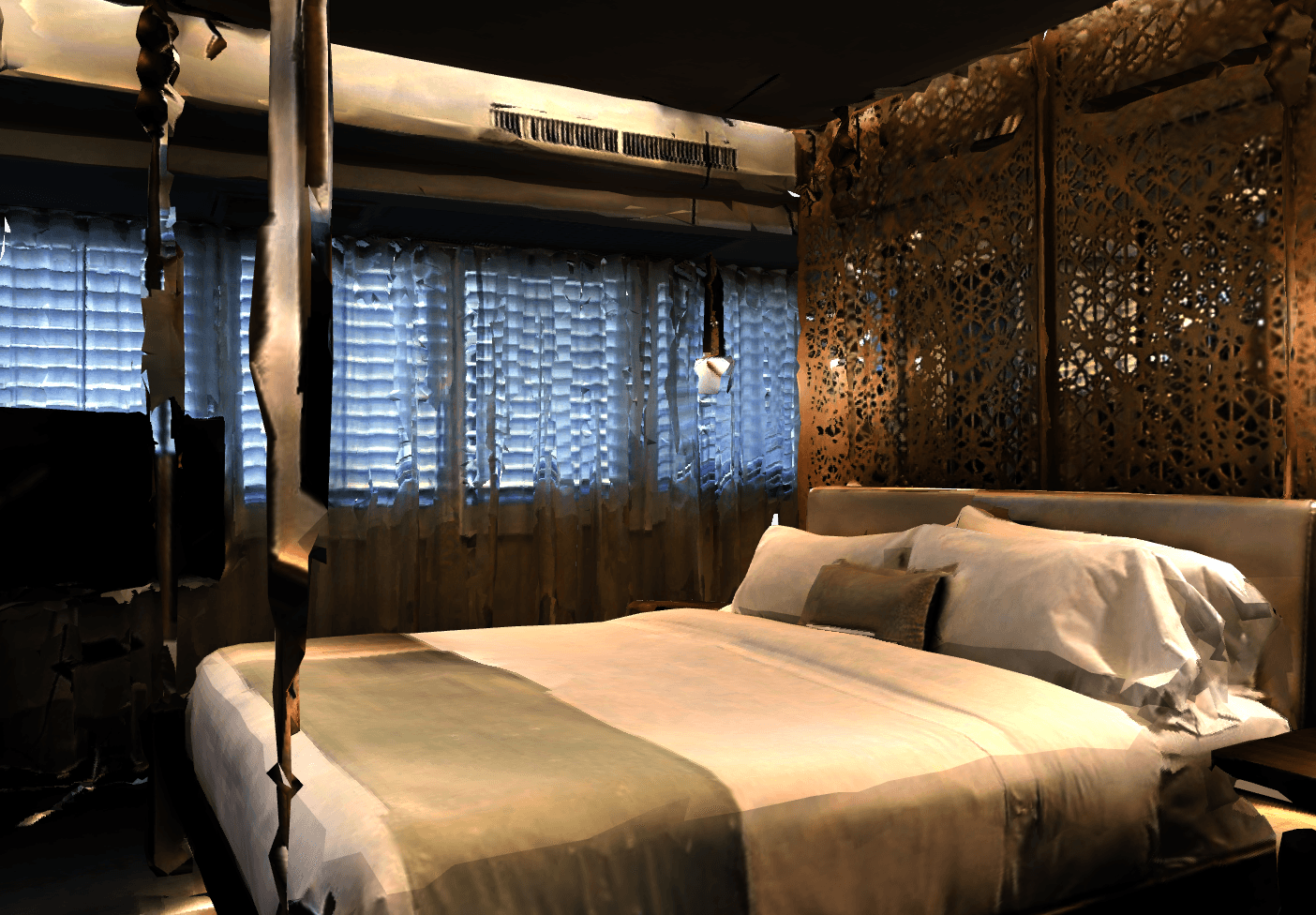}&
\includegraphics[width=.24\textwidth]{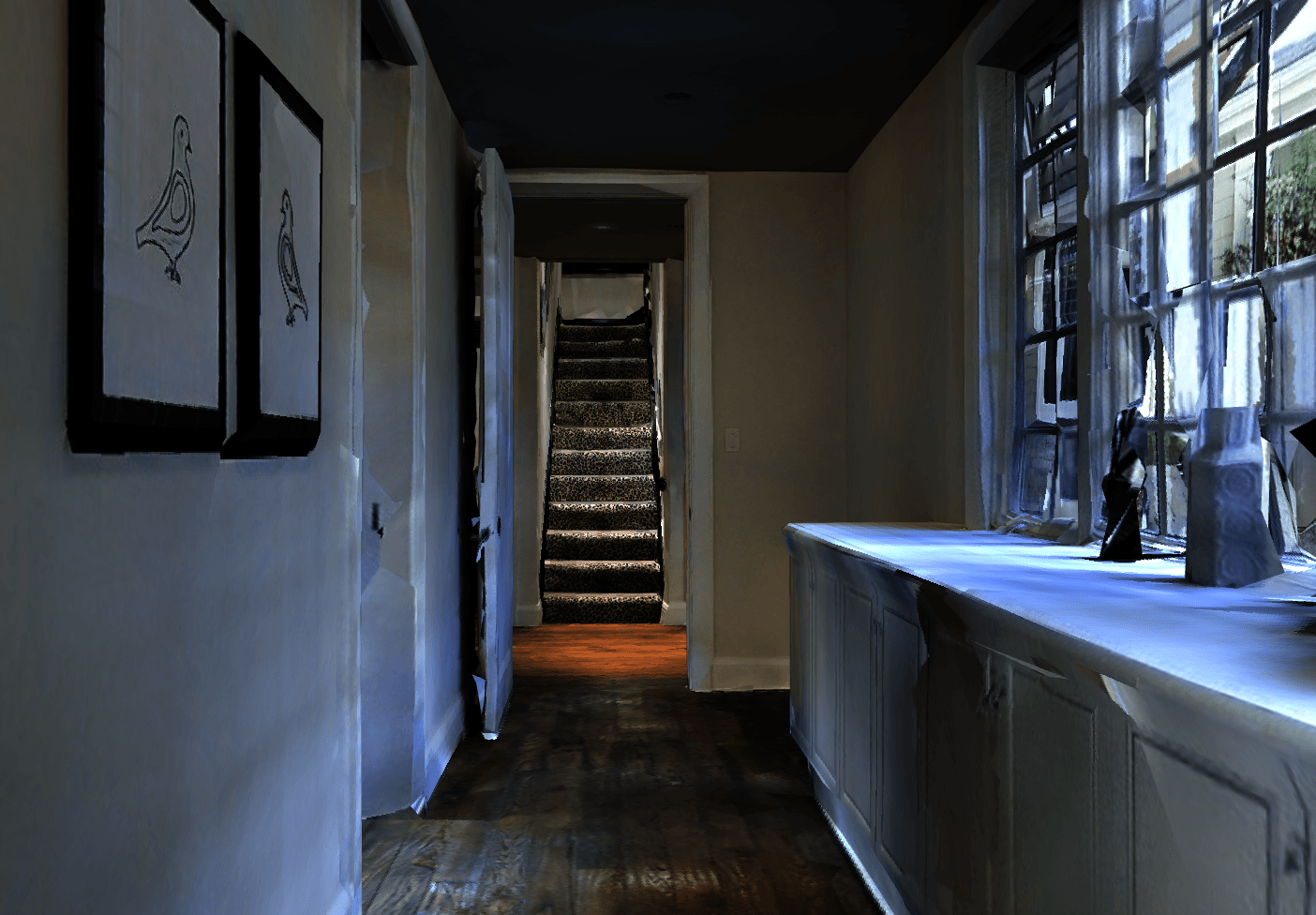}&
\includegraphics[width=.24\textwidth]{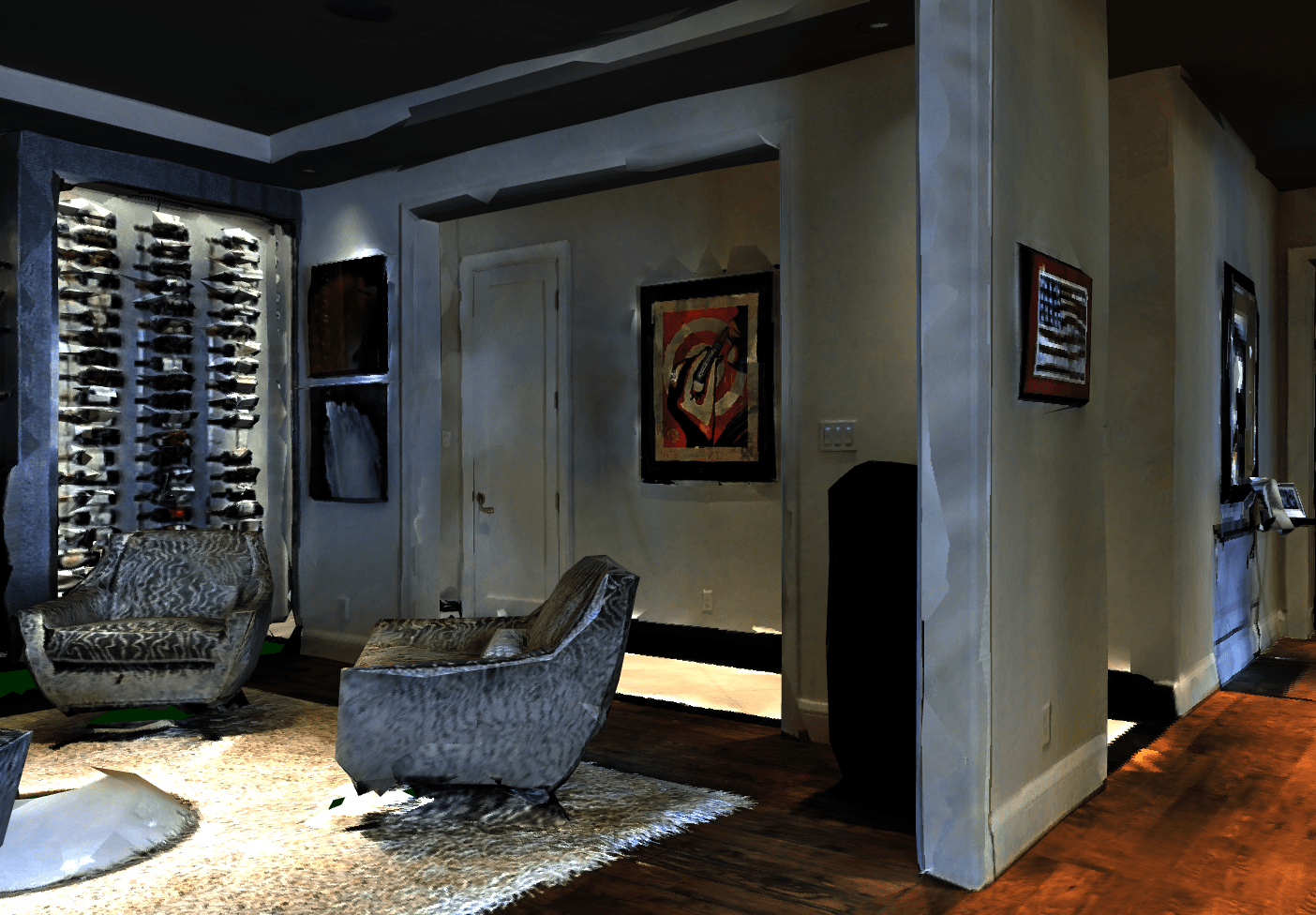}\vspace{-1mm}\\
\multicolumn{2}{c}{(c) 10-room M3D house} & \multicolumn{2}{c}{(d) 38-room M3D house}\\
\end{tabular}
\caption{Example houses from our indoor navigation datasets. For each of the four environments, we show an overhead view (top left), the same view with each room coded by a different color (top right), and two first-person views from within the environment (bottom).}
\label{fig:dataset}
\end{figure*}

%% file: tex/model.tex
\section{Methods}
\label{sec:methods}


We use MINOS to benchmark a set of recent navigation algorithms.
We assume that an agent interacts with the environment over discrete time steps in an episodic setup.
Each episode of interaction with the environment ends after a maximum number of time steps $T$.
At each time step $t$, the agent receives an observation $\oo_t$ and a scalar reward $r_t$ from the environment.
The observation $\oo_t = \tuple{\ss_t^1, \ldots, \ss_t^M}$ is a tuple consisting of $M$ raw sensory inputs $\ss_t^1, \ldots, \ss_t^M$ coming from different modalities.
Based on the observation, the agent takes an action $a_t$ from a discrete action set $\aA$ (we discretize the continuous action space provided by the simulator).

We study four end-to-end navigation algorithms.
The first three are based on asynchronous advantage actor-critic (A3C)~\citep{mnih2016asynchronous}.
The fourth is Direct Future Prediction~\citep{dosovitskiy2016learning}, which has shown good performance in a maze navigation task.
Note that since we consider agents acting in the continuous state space, we do not include the method of~\citet{gupta2017cognitive}, which assumes a discrete gridworld-like environment.
We now describe the methods in more detail.

\mypara{Feedforward A3C} is the most basic version of the asynchronous advantage actor-critic algorithm, where a feedforward convolutional network is used as a function approximator.
The agent is trained to estimate two quantities.
The first is the value function: the expected discounted sum of rewards from the current moment until the end of the episode.
The second is the policy: a distribution over the set of actions, indicating the degree of benefit expected from each action.
The value function is trained via the multi-step Bellman equation~-- a recurrent relation stating that the expected cumulative reward can be approximated as a sum of several rewards plus an agent's estimate at a future time step.
The policy is trained to maximize the probability of actions leading to larger-than-average rewards and to minimize the probability of actions leading to smaller-than-average rewards.
This is achieved via policy gradient with the value function serving as a baseline.
Further details are provided by~\citet{mnih2016asynchronous}.

\mypara{LSTM A3C} is an A3C agent in which the feedforward network is augmented by long short-term memory (LSTM) units, trained via backpropagation through time.
The vanilla A3C agent can only behave reactively based on the current observation; such an agent is unable to build an internal representation of the environment or execute temporally extended action sequences.
LSTM provides the agent with a simple memory, which can potentially help to alleviate these shortcomings of the feedforward agent.

\mypara{UNREAL} is a version of LSTM A3C that is augmented with auxiliary unsupervised tasks.
These extra tasks provide additional training signals to the network, leading to improved convergence and stability.
The auxiliary tasks include: value function replay, reward prediction, and ``pixel control''.
Further details are provided by~\citet{jaderberg2017reinforcement}.


\mypara{DFP} is the Direct Future Prediction algorithm~\citep{dosovitskiy2016learning}.
It differs from the aforementioned methods in that it does not explicitly aim to maximize future rewards. Instead it predicts future measurements~-- a set of low-dimensional sensory inputs.
The actions are then selected so as to maximize an objective function that is defined in terms of these measurements. The method can be seen as Monte Carlo reinforcement learning with a decomposed reward. Further details are provided by~\citet{dosovitskiy2016learning}.

%% file: tex/experiments.tex
\section{Experiments}

We use MINOS to evaluate the methods summarized in Section~\ref{sec:methods}. We compare the algorithms on goal-directed navigation, with goals specified either by their location relative to the agent, or by their semantic meaning.
We evaluate the effectiveness of different combinations of sensory modalities for navigation and measure the effect of environmental complexity on the methods' performance.
In contrast with many previous works, we do not evaluate the agents in the same environment they were trained in.
Rather, we study generalization to previously unseen environments.
Finally, we perform our experiments in both SUNCG environments and Matterport3D environments to investigate how algorithm performance is impacted by the domain difference between synthetic and reconstructed scenes.

\subsection{Experimental setup}

\subsubsection{Tasks}

Our goal-directed navigation tasks are set up as a sequence of trials.
In each trial, the agent is initialized at a random location in an indoor environment and has to reach a goal location.
We experiment with three ways of specifying goals: by their position relative to the agent (\emph{PointGoal}), or by the semantic category of the target object (\emph{ObjectGoal}) or target room region (\emph{RoomGoal}).
In this work, we specify the spatial goal using Euclidean distance and normalized direction to a randomly chosen point.  The room goal is specified as one of 9 distinct room classes (kitchen, bedroom, living room, toilet, bathroom, dining room, office, hallway, and miscellaneous), and the object goal is specified as an object class (we use doors as navigation targets in the reported experiments).

The agent is initialized at a random position and orientation in the free space of a house, and is provided with a target point, room, or object to which it must navigate, specified by the distance and direction towards the goal, or the semantic class of the room or object represented as a one-hot vector.
Each generated combination of start and goal positions is checked for navigability using a tile-based shortest-path computation.
The distance and direction measurements are the Euclidean distance to the goal point, or to the closest point on the goal object or room.
The trial ends once the agent reaches the goal, or after a fixed timeout of 500 steps (corresponding to $50\unit{s}$ of simulated time).  During training, the agent performs 10 trials in each environment, before moving on to a new randomly sampled environment.

\subsubsection{Environments}

We establish specific subsets of environments for benchmarking indoor navigation.
These were selected manually by verifying the realism and traversability of each environment floorplan.

From the SUNCG dataset we select a subset of $500$ single-floor houses of varying complexity, with one to ten rooms per house.
This dataset is split into $300/100/100$ training, validation, and test scenes.
The houses consist of a total of $2@737$ rooms, populated with $41@158$ object instances in a total floor area of approximately $110@000\unit{m^2}$.
On average, there are $5.5$ rooms per house with a mean floor area of $42\unit{m}^2$ per room.
Each house is populated with $82$ objects on average.
These houses represent a variety of environments including family homes, offices, and public spaces such as restaurants.

For the Matterport3D dataset, we adopt the training/validation/test split specified by the original dataset ($61/11/18$).
These environments comprise a total of $2@206$ room regions in $190$ floors.
On average, each house has $24.5$ rooms and a floor area of $560\unit{m}^2$, providing significantly larger interconnected environments for navigation.

For the SUNCG dataset, we create two variants of each house with different complexity: an \emph{empty} variant in which the scenes are emptied of furniture and include only architectural elements such as walls, ceilings, floors, doors, and windows, and a \emph{furnished} variant in which the scenes contain their full content except people and plants.

\subsubsection{Agent}
The agent is represented by a cylinder proxy geometry with a height of $1.09\unit{m}$ and a radius of $0.10\unit{m}$.
We use a continuous state space in our experiments. The motion of the agent is governed by simple rigid body physics.
The actions available to the agent include linear acceleration in forward or backward direction, as well as angular acceleration towards left or right.
We discretized the action space into \emph{turn left}, \emph{turn right}, and \emph{move forward} commands, which inject linear or angular acceleration, with linear acceleration of $20\unit{m~s^{\text{-}2}}$, angular acceleration of $4\pi\unit{rad~s^{\text{-}2}}$ (clockwise or counterclockwise), and maximum speeds of $2\unit{m~s^{\text{-}1}}$ and $4\pi\unit{rad~s^{\text{-}1}}$, respectively.
These settings produce linear steps of about $20\unit{cm}$ and turns of about $23\degree$.

For the multimodal agent experiments, the agent is provided with combinations of vision, depth, and contact sensors in addition to the goal signal.
Vision is provided by a singe grayscale camera located at height $1.09\unit{m}$ and configured with a field of view of $90\degree$.
Images are fed to agents at $84\timess84$ pixel resolution.
The depth sensor is co-located with the vision sensor, and outputs ground-truth depth in the $[0,10\unit{m}]$ range, quantized to byte precision with no noise.
Four contact-force sensors are placed at height $0.3\unit{m}$ above ground on the surface of the agent cylinder proxy geometry, oriented in the four cardinal directions with respect to the agent.
The contact sensor configuration encodes collision impulse responses as a binary ``contact'' signal in each direction.

\subsubsection{Training details}
The agents are trained and tested over episodes lasting up to $500$ time steps, with $10$ steps per second of simulated time.
Each agent is trained for a total of $13.2\unit{M}$ time steps, corresponding to roughly 15 days of experience.
Average training speed with four simulation threads is about $167$ steps per second, amounting to approximately $14.4\unit{M}$ steps per day.
We run four such training processes (four simulation threads each) on a single Nvidia Titan X Pascal GPU, yielding a total of $57.6\unit{M}$ steps per day ($668$ steps per second).

We use an epsilon-greedy random exploration schedule, starting with a fully random policy and decaying to approximately 10\% probability of random actions by the end of training.
The navigation goal is chosen at random for each episode.
The DFP agents are trained against a future prediction loss with measurements at time $t$: ${\mm_t = \tuple{d_t, x_t, z_t, t}}$, where $d_t$ is the Euclidean distance to the goal and $(x_t, z_t)$ is the normalized 2D direction to the goal for the PointGoal task, or the dot product of a one-hot representation of the room category in the RoomGoal task with the goal room category.
On-policy actions are chosen by selecting the action that minimizes an objective with a linear combination of the predicted distance $d_t$ and the normalized time $t$ with equal weight.
The temporal offsets $\{\tau_1, \ldots, \tau_n\}$ are set to $1$, $2$, $4$, $8$, $16$, and $32$ steps.
The A3C FF, A3C-LSTM and UNREAL agents are all trained under a reward function computing the difference in Euclidean distance to the goal $d_t$ and normalized time $t$ for each time step (matching the objective used for the DFP agent).
We use the same training hyperparameters as reported by \citet{jaderberg2017reinforcement}, but with only four asynchronous threads.
For the RoomGoal task, we provide the one-hot goal room category vector as an additional input concatenated with the agent state.

\subsection{Results}

At test time, agents are tested for $10$ episodes per scene, in a fixed permuted order of scenes with a set of pre-sampled starting configurations selected to span a range of distances from the goal.
Agent performance is evaluated by the overall episode \emph{success} rate (fraction of episodes ending with the agent arriving at the goal), reported as percentage averaged over all testing episodes.

\Cref{tab:results_complexity} shows the performance of the various navigation agents on goal-directed navigation with variable goal specification (PointGoal or RoomGoal), environment complexity (size, presence of furniture), and environment realism (synthetic or reconstructed).
We now analyze these results.

\input{figures_tex/tab_results_complexity.tex}

\paragraph{Relative performance of the agents.}
On most tasks, the UNREAL agent performs best, followed by DFP and A3C-LSTM.
A3C-FF failed to learn any meaningful policy, and performs worse than a random agent in some cases, which could be due to hyperparameter selection.
Despite the lack of memory, DFP outperforms A3C-LSTM and UNREAL for the smaller and less cluttered SUNCG environments in point navigation.
In more challenging setups using larger Matterport3D environments and for the semantic room navigation task, UNREAL significantly outperforms the other approaches.
This is likely due to the combination of memory and supervision through auxiliary learning.

\paragraph{Environment complexity.}
The performance of all methods declines significantly in large and cluttered environments.
The best-performing agent in the PointGoal task is successful in about $80\%$ of trials in the simplest two-room empty SUNCG environments.
However, in the most complex PointGoal setup~-- Matterport 3D houses with up to 24 rooms~-- all agents have success rate of $20\%$ or lower.
These results indicate that even in the simplest scenario the performance is not perfect, and that existing RL methods fail in large, cluttered, realistic environments.

\paragraph{Spatial and semantic goals.}
The RoomGoal task is more difficult for all algorithms than PointGoal.
This is likely due to the sparsity of the RoomGoal reward signal, which only indicates whether the agent is in a room with a type matching the goal room type.

\subsection{Navigation with multimodal sensory input}

In the previous experiments, all agents were navigating based solely on visual input.
We now compare agents equipped with different sensor suites to investigate the impact of multimodal sensory input on navigation performance in an ObjectGoal task where the navigation target is a randomly chosen door.
We adapt the visual DFP agent by providing alternative or additional modalities as input to create several multimodal agents: target distance and direction measurements only; vision only; depth only; contact force only; vision and contact force; vision, depth, and contact force.
As in prior experiments, agents are trained for $13.2\unit{M}$ steps.
Agent performance is evaluated by the success rate, and by a \emph{speed} measure which is the fraction of time left at the end of the episode (for all episodes).

\Cref{tab:average_steps} reports the performance of each agent after training on several sets of novel environments with different complexity: empty single-room SUNCG environments, all empty SUNCG houses, furnished single-room SUNCG environments, and all furnished SUNCG houses. In the simplest setting of empty single-room environments, all agents do well. In particular, the combination of measurements and contact force input performs best, likely due to the simplicity and sufficiency of the goal direction and contact signals for navigating towards a single target door (usually unobstructed in this trivial setting). For empty multi-room houses (up to ten rooms each), the depth modality performs particularly well, whereas the vision modality does not increase performance significantly. This is likely due to the fact that in the absence of clutter, the additional benefit of visual input is limited. The full multimodal agent outperforms the ablated versions in this setting. The full multimodal agent likewise performs best in the furnished settings. Among individual modalities in the furnished settings, depth confers the strongest advantage.

\input{figures_tex/tab_results.tex}

%% file: figures_tex/tab_results_complexity.tex

\begin{table*}
	\centering\small
\resizebox{\textwidth}{!}{
	\begin{tabular}{@{}ccccccccc@{}}
	\toprule
	 \multicolumn{4}{c}{Environment}  & \multicolumn{5}{c}{Agent}\\
	 \cmidrule(lr){1-4} \cmidrule(lr){5-9}
    Task      & Dataset      & Clutter   & Size  & Random & A3C-FF & A3C-LSTM & DFP & UNREAL \\
    \midrule
    PointGoal & SUNCG        & Empty     & Small & 23.8 & 10.1 & 69.1 & \textbf{80.3} &         72.9 \\
    PointGoal & SUNCG        & Empty     & Medium &  8.6 &  7.2 & 57.4 & \textbf{64.1} &         63.2 \\
    PointGoal & SUNCG        & Furnished & Small &  9.5 & 16.1 & 60.9 & \textbf{64.5} &         64.1 \\
    PointGoal & SUNCG        & Furnished & Medium &  6.3 &  7.9 & 41.3 &       43.6  & \textbf{45.3} \\
    PointGoal & Matterport3D & Furnished & Small &  0.0 &  2.0 & 32.0 &       27.3  & \textbf{38.0} \\
    PointGoal & Matterport3D & Furnished & Medium &  0.0 &  2.0 & 0.0 &       18.2  & \textbf{20.0} \\
    RoomGoal  & SUNCG        & Furnished & Small & 10.0 & 25.7 & 30.0 &       22.5  & \textbf{58.6} \\
    RoomGoal  & SUNCG        & Furnished & Medium &  3.1 &  6.9 &  7.2 &        4.0  & \textbf{32.0} \\
    RoomGoal  & Matterport3D & Furnished & Small &  5.0 & 12.0 & \textbf{14.0} &       13.6  & \textbf{14.0} \\
	\bottomrule
	\end{tabular}
}
	\caption{
	Average episode success rate for agents trained on PointGoal and RoomGoal tasks, tested on novel environments of varying complexity. SUNCG `Small' refers to two-room houses while SUNCG `Medium' contains three-to-five-room houses. Matterport3D `Small' contains environments with up to 10 rooms, and Matterport3D `Medium' refers to environments with up to 24 rooms. Note that all agents exhibit significant performance degradation as the size and complexity of the environment increases.}
	\label{tab:results_complexity}
\end{table*}

%% file: figures_tex/tab_results.tex
\begin{table*}[t]
	\centering
	\begin{tabular}{@{}lcccccccc@{}}
	\toprule
	 & \multicolumn{2}{c}{empty room} & \multicolumn{2}{c}{empty house} & \multicolumn{2}{c}{furnished room} & \multicolumn{2}{c}{furnished house} \\
	\cmidrule(lr){2-3} \cmidrule(lr){4-5} \cmidrule(lr){6-7} \cmidrule(lr){8-9}
	modalities & success & speed & success & speed & success & speed & success & speed \\
	\midrule
	None & 23.1 & 12.7 & 18.8 & 11.3 & 9.5 & 5.2 & 6.5 & 3.1 \\
	\textbf{M}easurements & 90.6 & 68.1 & 66.5 & 49.5 & 46.2 & 27.2 & 25.8 & 12.2 \\
	\textbf{M}+\textbf{V}ision & 86.8 & 71.2 & 68.7 & 53.7 & 48.4 & 38.6 & 30.4 & 19.6 \\
	\textbf{M}+\textbf{D}epth & 92.7 & 77.6 & 74.8 & 63.7 & 64.2 & 51.7 & 43.7 & 28.5 \\
	\textbf{M}+\textbf{F}orces & \textbf{96.7} & 78.7 & 70.3 & 56.3 & 56.0 & 37.8 & 39.9 & 29.1 \\
	\textbf{M}+\textbf{V}+\textbf{F} & 94.8 & \textbf{80.4} & 74.3 & 61.1 & 54.4 & 41.7 & 41.8 & 29.3 \\
	\textbf{M}+\textbf{D}+\textbf{V}+\textbf{F} & 94.4 & 80.0 & \textbf{78.1} & \textbf{65.3} & \textbf{64.7} & \textbf{52.4} & \textbf{48.0} & \textbf{35.0} \\
	Human & 100.0 & 92.1 & 100.0 & 87.7 & 100.0 & 90.5 & 99.0 & 84.9 \\
	\bottomrule
	\end{tabular}
	\caption{Performance of trained multimodal DFP agents on novel SUNCG environments. `None' is a random policy (no perceptual input from the environment), `Human' reports human performance on one test episode of average difficulty per scene. The other rows report the performance of agents equipped with different sets of perceptual modalities. Each pair of columns reports results in a different setting. From left to right: empty single-room environments, empty houses (all), furnished single-room environments, furnished houses (all). Note that the single-room environments are less challenging than the small two-room environments in Table~\ref{tab:results_complexity}.}
	\label{tab:average_steps}
\end{table*}

%% file: tex/discussion.tex
\section{Discussion}

We presented a multimodal simulation platform that is designed to support the development of multisensory models for goal-directed navigation in indoor environments. Our simulator provides a suite of sensory input modules that can be flexibly combined. By leveraging two large-scale datasets of indoor environments and augmenting the data through controlled variation of appearance and clutter, we provide orders of magnitude more indoor environments for training and testing than previously available. Our experiments demonstrate that current deep reinforcement learning approaches fail in large, realistic indoor environments, and that multimodality is beneficial in learning to act in cluttered indoor scenes.
